\documentclass[runningheads]{llncs}
\usepackage[T1]{fontenc}
\usepackage{graphicx}
\usepackage{amsmath}
\usepackage{amssymb}
\usepackage{amsfonts}
\usepackage{placeins}
\usepackage[hidelinks]{hyperref}

\begin{document}
\title{Unified Multimodal Brain Decoding via Cross-Subject Soft-ROI Fusion}
%
%

\author{Xuanyu Hu\thanks{Corresponding author.}\inst{1}}
\authorrunning{X. Hu}

\institute{The University of Manchester, Manchester, UK\\
\email{huxuanyu0523@126.com}}

\maketitle              
\begin{abstract}
Multimodal brain decoding aims to reconstruct semantic information that is consistent with visual stimuli from brain activity signals such as fMRI, and then generate readable natural language descriptions. However, multimodal brain decoding still faces key challenges in cross-subject generalization and interpretability. We propose a BrainROI model and achieve leading-level results in brain-captioning evaluation on the Natural Scenes Dataset (NSD) dataset. Under the cross-subject setting, compared with recent state-of-the-art methods and representative baselines, metrics such as BLEU-4 and CIDEr show clear improvements. Firstly, to address the heterogeneity of functional brain topology across subjects, we design a new fMRI encoder. We use multi-atlas soft functional parcellations (soft-ROI) as a shared space. We extend the discrete Regions of Interest (ROI) Concatenation strategy in MINDLLM to a voxel-wise gated fusion mechanism (Voxel-gate). We also ensure consistent ROI mapping through global label alignment, which enhances cross-subject transferability. Secondly, to overcome the limitations of manual and black-box prompting methods in stability and transparency, we introduce an interpretable prompt optimization process. In a small-sample closed loop, we use a locally deployed Qwen model to iteratively generate and select human-readable prompts. This process improves the stability of prompt design and preserves an auditable optimization trajectory. Finally, we impose parameterized decoding constraints during inference to further improve the stability and quality of the generated descriptions.


\end{abstract}

\section{Introduction}

Non-invasive neuroimaging (e.g., fMRI) provides observable neural representations for studying human visual cognition. With the development of vision-language large models (VLMs/MLLMs), recent methods attempt to map brain activity into a visual semantic feature space and perform decoding tasks such as image captioning, localization, and question answering on frozen multimodal language models. A representative work, UMBRAE~\cite{umbrae2024}, adopts the "fMRI → visual space → frozen MLLM" paradigm, demonstrating the potential of unified zero-shot multi-task decoding while avoiding the cost of end-to-end large-model fine-tuning. However, UMBRAE still leaves room for improvement in cross-subject generalization: to deal with inter-subject variability, it introduces subject-specific tokenizers, which make the model contain subject-dependent parameters, hindering the realization of a fully parameter-shared, subject-agnostic solution.

To improve cross-subject robustness, MINDLLM~\cite{mindllm2025} explores using neuroscience priors, namely functional brain regions of interest (ROIs), as shared anchors across subjects. This design aims to reduce representation instability caused by the spatial sparsity of fMRI signals and inter-subject variability. However, MINDLLM relies on a discrete feature concatenation strategy. This strategy makes it hard to achieve an adaptive trade-off between coarse ROI priors and fine-grained voxel-level information. As a result, the question remains open: how can we build a more effective unified mechanism that balances a transferable shared space with detailed voxel-level representations? Meanwhile, IPO~\cite{ipo2024} introduces interpretable prompt optimization for MLLM tasks. IPO treats an LLM as an "optimizer". The method iteratively generates human-readable candidate prompts in a small-sample closed loop. The method then selects the best prompt using target-task metrics. This design offers both strong performance and interpretability. However, existing work has not systematically integrated this idea into the fMRI→MLLM brain-decoding pipeline. Finally, the quality of the generated text is also crucial. Brain-captioning benchmarks mainly rely on metrics such as BLEU and CIDEr. If a model's generation distribution shows systematic bias relative to the references, the scores can drop sharply. Common biases include outputs that are too short or too long, and repeated n-grams. These issues can hurt both automatic metrics and human readability, even when the overall meaning is roughly correct.

To address these three challenges (cross-subject generalization, prompt stability and interpretability, and generation consistency), we propose a BrainROI model. On the encoding side, we construct a multi-atlas soft-ROI shared space and enforce ROI semantic consistency across subjects and atlases via global label alignment; we further introduce a voxel-wise gated fusion module (Voxel-gate) to enable interpretable, adaptive integration of multi-atlas priors. On the prompting side, we adopt an interpretable prompt optimization strategy, using a locally deployed large language model to automatically generate, evaluate, and select human-readable prompts while retaining an auditable optimization trace. On the generation side, we employ parameterized constrained decoding to enhance output stability and alignment with reference captions. We systematically evaluate the proposed approach under the brain captioning protocol on the NSD dataset. Experimental results show consistent improvements over UMBRAE, MINDLLM, and other methods across multiple metrics, and ablation studies further verify the contribution of each component to cross-subject robustness and reproducibility.

Our contributions are summarized as follows:
\begin{enumerate}
    \item We propose a cross-subject shared space based on multi-atlas soft ROIs, together with global label alignment. We also introduce Voxel-gate. This design balances transferability and fine-grained spatial representation.

    \item We integrate an interpretable prompt optimization system into the fMRI decoding pipeline. The system uses a locally deployed LLM to generate and select human-readable prompts. The system also keeps an auditable optimization trace, which improves stability and interpretability.

    \item We propose a reference-consistency–oriented decoding strategy with constrained generation. This strategy improves text readability and evaluation scores.

    \item We provide systematic empirical results under the NSD brain captioning protocol. We also release the scripts and hyperparameter configurations needed for reproduction. Our code is available at: https://github.com/xuanyu-op/bd-Soft-ROI-brain-decoding
\end{enumerate}

\section{Related Work}
\subsection{Unified Multimodal Brain Decoding}
Recent years have made clear progress in decoding natural language from fMRI signals. This progress benefits from the rich semantic space provided by large pretrained vision–language models. A primary approach is to train an fMRI encoder which aligns neural responses to a shared representation space, such as CLIP~\cite{clip2021}. A language model then generates text from the aligned features. This pipeline effectively maps high-dimensional brain activity into semantic vectors that pretrained models can use. In addition, research on non-invasive continuous language reconstruction shows that we can recover a coherent stream of language from dynamic brain activity, instead of only generating label-like captions for single images ~\cite{tang2023semantic}. Researchers have also pushed toward "unified" decoding ability. Later methods extend from a single text-generation task to multiple downstream tasks. For example, OneLLM~\cite{onellm2024} aligns multiple modalities, including fMRI, within one framework to support multi-task learning. UMBRAE connects an fMRI encoder to a frozen multimodal large language model. It maps brain signals to proxy image features and enables zero-shot inference on tasks such as image captioning and visual question answering. MINDLLM proposes an end-to-end, subject-agnostic fMRI-to-text system that uses "brain instruction tuning" to enhance memory and reasoning. Overall, research in this area is moving from single-output decoding to unified brain decoding that supports multiple, more interactive tasks. However, major challenges remain: cross-subject variability is high, data are limited, and the signal-to-noise ratio is often low. Under these conditions, it is still difficult to achieve both stable generation quality and an interpretable, unified decoding capability.

\subsection{Cross-subject Generalization}
Due to individuals differing greatly in functional topology and anatomical structure, cross-subject generalization has long been a core challenge in fMRI decoding. Early studies often trained a separate model for each subject. This approach is expensive and hard to scale \cite{mozafari2020bigbigan,takagi2023ldm_cvpr}. To address this issue, recent work has explored several unified modeling strategies. MindBridge~\cite{mindbridge2024} uses adaptive pooling to map a variable number of voxels into a fixed-length input, which enables cross-subject training. UniBrain~\cite{unibrain2024} groups and samples voxels to fit Transformer-based models. UMBRAE adopts a hybrid design that combines subject-specific tokenizers with a shared encoder. It first applies light subject-level normalization, and it then shares the encoder to learn common patterns. MINDLLM proposes a neuroscience-informed attention mechanism. It uses predefined ROI atlases as priors. It builds attention keys from ROI embeddings together with 3D voxel coordinates. This design promotes cross-subject alignment through shared functional organization. It is also worth noting that MINDLLM fuses ROI information through discrete feature concatenation. This design may need a finer-grained weighting mechanism when it faces cases such as inconsistent granularity across multiple atlases. For this reason, we further study how to achieve adaptive and interpretable fusion across multi-atlas priors. Our goal is to strengthen cross-subject representations and improve transfer performance.

\subsection{Prompt Optimization for MLLM}
The performance of large multimodal models is highly sensitive to prompts. Early methods relied on manually designed discrete prompts. This approach is intuitive, but it is time-consuming and often suboptimal. Later work introduced gradient-based tuning of continuous or prefix prompts, such as soft prompts, Prefix-Tuning, and CoOp \cite{prefixtuning2021,l2p2022,coop2022,vpt2022}. These methods can achieve strong performance, but they offer limited interpretability. They can also overfit easily in low-data settings.

More recently, interpretable prompt optimization has emerged. This line of work treats a strong LLM as an "optimizer" \cite{llm_prompt_engineer_2023,apo2023}. The LLM iteratively generates and filters readable prompts based on task feedback. This design balances automation and transparency. It can also generalize better in some cases, as shown by methods such as IPO. To the best of our knowledge, this idea has not been systematically applied to brain-decoding tasks. Therefore, we introduce interpretable prompt optimization into a unified brain-decoding pipeline. Our goal is to automatically discover effective and auditable prompts in a small-sample closed loop and to improve the quality and stability of downstream text generation.

\section{Method}

\subsection{Overall Framework}
\begin{figure}
    \centering
    \includegraphics[width=1\linewidth]{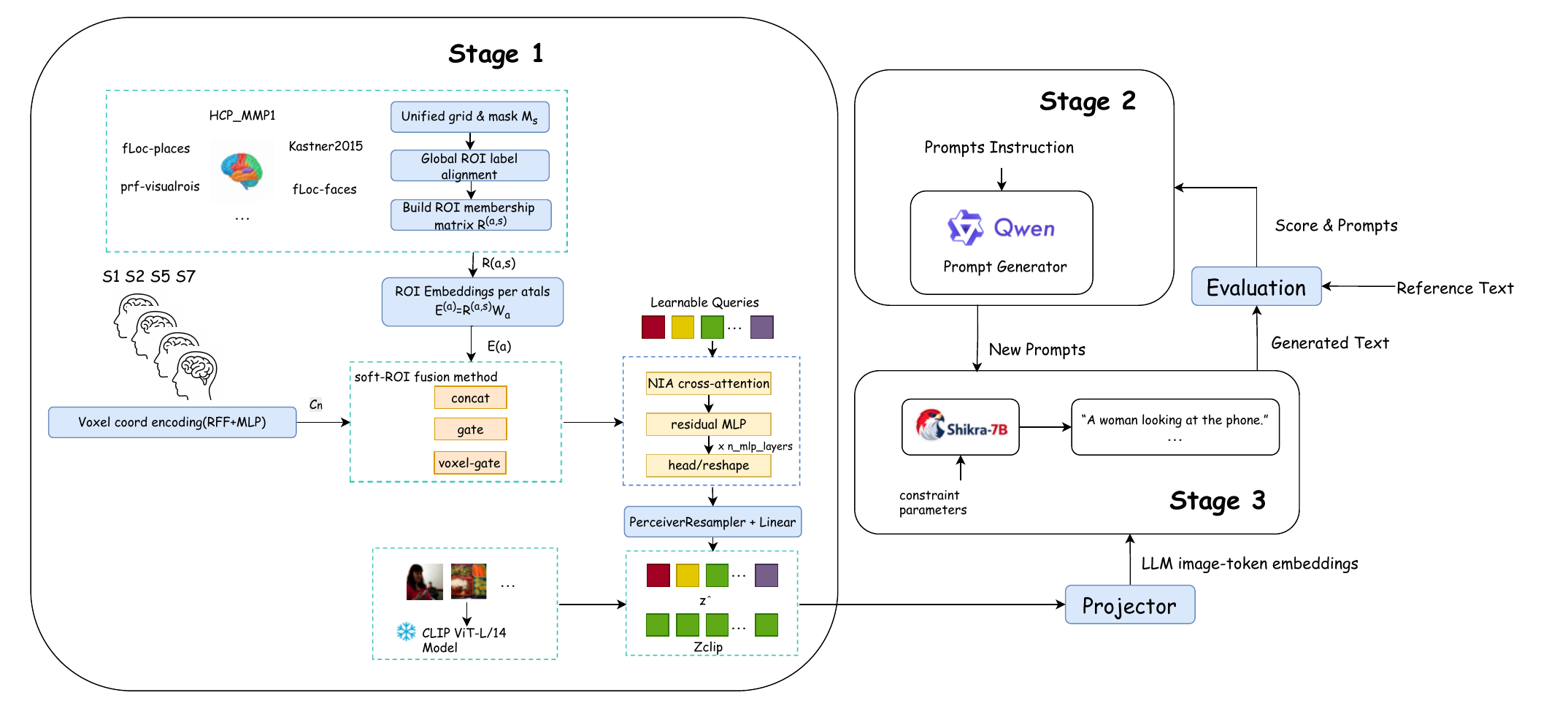}
    \caption{Overall framework of our three-stage brain-decoding pipeline. The framework illustrates the integration of soft-ROI fusion, Voxel-gate mechanism, and interpretable prompt optimization.}
    \label{fig:overall_framework}
\end{figure}

Our brain-decoding framework follows a three-stage pipeline as illustrated in Fig.~\ref{fig:overall_framework}: First, we design an fMRI encoder based on multi-atlas (soft-ROI) with gated fusion. This encoder enables efficient and robust cross-subject feature extraction. Next, we introduce an LLM-based closed-loop optimization process. This process automatically generates and selects optimal human-readable prompts. Finally, we feed the encoded brain-activity features and the optimized prompts into a frozen MLLM with parameterized constraints, and the model generates high-quality descriptive text.

\textbf{Stage 1: Cross-subject fMRI Encoding and Alignment to the CLIP Space}

This stage aims to encode fMRI signals from different subjects, with varying numbers of voxels, into a fixed-length sequence of visual tokens, and to align them with the intermediate-layer patch feature space of a CLIP image encoder.

We first perform global ROI label alignment under a unified voxel grid and the subject-specific mask $M_s$, mapping ROIs from different subjects and different brain atlases into a shared column space with a fixed ordering, and constructing the ROI membership matrix $R^{(a,s)}$ for each atlas.

Next, we map $R^{(a,s)}$ to an ROI prior embedding $E^{(a)} = R^{(a,s)} W_a$, and provide three switchable fusion strategies along the atlas dimension: Concatenation, Gate (voxel-independent global weighting with $\alpha_a$), and Voxel-gate (with $\alpha_{n,a}$).

We then combine the fused prior $\tilde{E}_n$ with the voxel coordinate encoding $c_n$ to construct the key vector $k_n$, project the voxel-wise fMRI signal $x_n$ into the value vector $v_n$, and aggregate voxel-level $(K,V)$ via Neural Information Aggregation cross-attention with learnable queries. The resulting representation is compressed by a PerceiverResampler into fixed-length visual tokens $Z_{\text{fMRI}} \in \mathbb{R}^{L \times D_{\text{CLIP}}}$. Finally, given the same stimulus image, we extract $Z_{\text{CLIP}}$ andalign the two using an mean squared error objective:$\mathcal{L}_{\text{align}} = \left\| Z_{\text{fMRI}} - Z_{\text{CLIP}} \right\|_2^2 .$

\textbf{Stage 2: Interpretable Prompt Optimization}

To replace handcrafted prompts and unreadable soft prefix prompts, we adopt an IPO-style few-shot closed loop. We use a locally deployed Qwen~\cite{qwen25_2024} model as the prompt generator. In each iteration, it produces several new human-readable candidate prompts according to the Prompts Instruction, and these candidates are fed into the Stage 3 generation-and-evaluation pipeline for scoring and ranking. Meanwhile, we keep the full evolution trace of "prompt–score" pairs to improve auditability and transparency. 

\textbf{Stage 3: Projector Injection and Parameterized Constrained Generation}

At inference, the visual tokens $Z_{\text{fMRI}}$ from Stage 1 are mapped by a Projector into image-token embeddings that can be injected into the MLLM, and are then jointly provided with the best prompt from Stage 2 to the frozen Shikra-7B~\cite{shikra2023} to generate caption text. In addition, we introduce parameterized decoding constraints during generation to enhance output stability and consistency with the reference distribution.

By combining the above three components, our framework achieves robust cross-subject performance and high-quality text generation.

\subsection{Cross-subject Soft-ROI gated Fusion Encoder}
\subsubsection{ROI Membership Indicator Matrix and Global Label Alignment}

To enable the encoder to leverage neuroscience priors while remaining comparable across subjects, we construct an ROI membership indicator matrix for each subject $s$ and each atlas $a$, $R^{(a,s)} \in \mathbb{R}^{N_s \times K_a}$, where rows correspond to voxels inside the subject mask, and columns follow a globally aligned ROI label order for atlas $a$.

\paragraph{Step 1: Spatial reference alignment (unified grid and mask).}
We use each subject's functional volume nsdgeneral.nii.gz as the reference grid for that subject. We then resample all atlases to this grid using nearest-neighbor interpolation. This operation removes differences in voxel coordinate systems while keeping discrete label semantics unchanged. We define the subject mask $M_s$ as the set of voxels whose values equal 1 in nsdgeneral.nii.gz. We extract voxel coordinates inside the mask in a fixed $(i,j,k)$ order and store them in voxel\_indices.npy. This file serves as the row-index reference and ensures that each row of every $R^{(a,s)}$ corresponds to a unique voxel location.

\paragraph{Step 2: Semantic reference alignment (global ROI label space).}
For each atlas $a$, we compute the union of all non-zero ROI labels within the masks of all subjects to form a global label set, and fix its ordering as $\{t_1, t_2, \dots, t_{K_a}\}$, thereby obtaining a cross-subject consistent "shared column space". This constraint ensures that, along the column dimension, $R^{(a,s)}$ for any subject expresses the same set of ROI semantics, making representations comparable across subjects.

\paragraph{Step 3: Construction of the ROI membership matrix $R^{(a,s)}$.}
Let the integer label of the $i$-th in-mask voxel be $\ell_i \in \{0\} \cup \mathcal{L}_a$, where $0$ denotes background and $\mathcal{L}_a$ is the set of valid ROI labels in atlas $a$. From Step~2, we obtain the globally aligned label sequence $label\_ids\_global^{(a)} = \{t_1, \ldots, t_{K_a}\}$. For the row vector $r_i \in \mathbb{R}^{K_a}$, we define it as a one-hot indicator vector over the globally aligned label sequence: for each $j=1,\ldots,K_a$, the $j$-th entry of $r_i$ is set to $1$ if and only if $\ell_i=t_j$, and set to $0$ otherwise. If $\ell_i=0$ or $\ell_i \notin \mathcal{L}_a$, we set $r_i=0$. We then get $R^{(a,s)} = [r_1; \ldots; r_{N_s}] \in \mathbb{R}^{N_s \times K_a}$. Each row of $R^{(a,s)}$ matches a fixed voxel position defined by voxel\_indices.npy, and each column corresponds to one element in the global label sequence. This design enforces strict comparability across subjects in both semantic and spatial dimensions.

\subsubsection{Voxel Coordinate Encoding and ROI Embeddings}

After aligning the row and column semantics of $R^{(a,s)}$, we separately vectorize voxel coordinates and ROI memberships, obtaining the spatial prior features $c_n$ and the functional prior features $E_n^{(a)}$, which are used for subsequent key construction and attention aggregation.
\begin{enumerate}
    
\item\textbf{Voxel coordinate encoding:} We normalize voxel coordinates within the mask (mapping them to $[-1,1]^3$), and then project them to $d_c$ dimensions using Random Fourier Features (RFF) followed by a multi-layer perceptron (MLP), yielding a coordinate representation for each voxel $c_n \in \mathbb{R}^{d_c}$. This representation preserves spatial discriminability while providing a positional reference for cross-subject shared encoding.

\item\textbf{ROI linear embedding:} For each atlas $a$, we learn a cross-subject shared linear embedding matrix $W_a \in \mathbb{R}^{K_a \times d_{\text{roi}}}$, and map $R^{(a,s)}$ to an ROI prior embedding$E^{(a)} = R^{(a,s)} W_a \in \mathbb{R}^{N_s \times d_{\text{roi}}}$. This design decouples shared atlas semantic parameters from subject-specific spatial distributions: $W_a$ captures atlas-level functional semantics, while $R^{(a,s)}$ provides a subject's ROI distribution under the unified shared column space. Even if some global ROI labels do not appear within a given subject's mask, the corresponding columns can remain zero, without affecting cross-subject consistency in shape and semantics.
\end{enumerate}

\subsubsection{Key Construction}

Within the shared column space, we fuse the coordinate encoding $c_n \in \mathbb{R}^{d_c}$ and the multi-atlas ROI embeddings $\{E^{(a)}_n \in \mathbb{R}^{d_{\text{roi}}}\}_{a=1}^{A}$ into an attention key vector $k_n \in \mathbb{R}^{d_k}$. We provide three switchable strategies: Concatenation, Gate, and Voxel-gate. All strategies produce keys with the same output shape $K \in \mathbb{R}^{N_s \times d_k}$. We set the coordinate encoding dimension to match the key dimension, i.e., $d_c = d_k$.

\paragraph{(i) Concatenation.}

In the Concatenation strategy, we directly concatenate the coordinate encoding and the ROI embeddings from each atlas along the feature dimension:
$z_n = [c_n; E_n^{(1)}; \dots; E_n^{(A)}], k_n = W_k z_n$, where $W_k$ is a linear projection matrix. This strategy is the simplest, but the contribution of each atlas is not separable, and it lacks an adaptive trade-off across different voxel locations.

\paragraph{(ii) {Gate}: atlas-level gated weighting.}

\begin{figure}
    \centering
    \includegraphics[width=\linewidth,height=0.32\textheight,keepaspectratio]{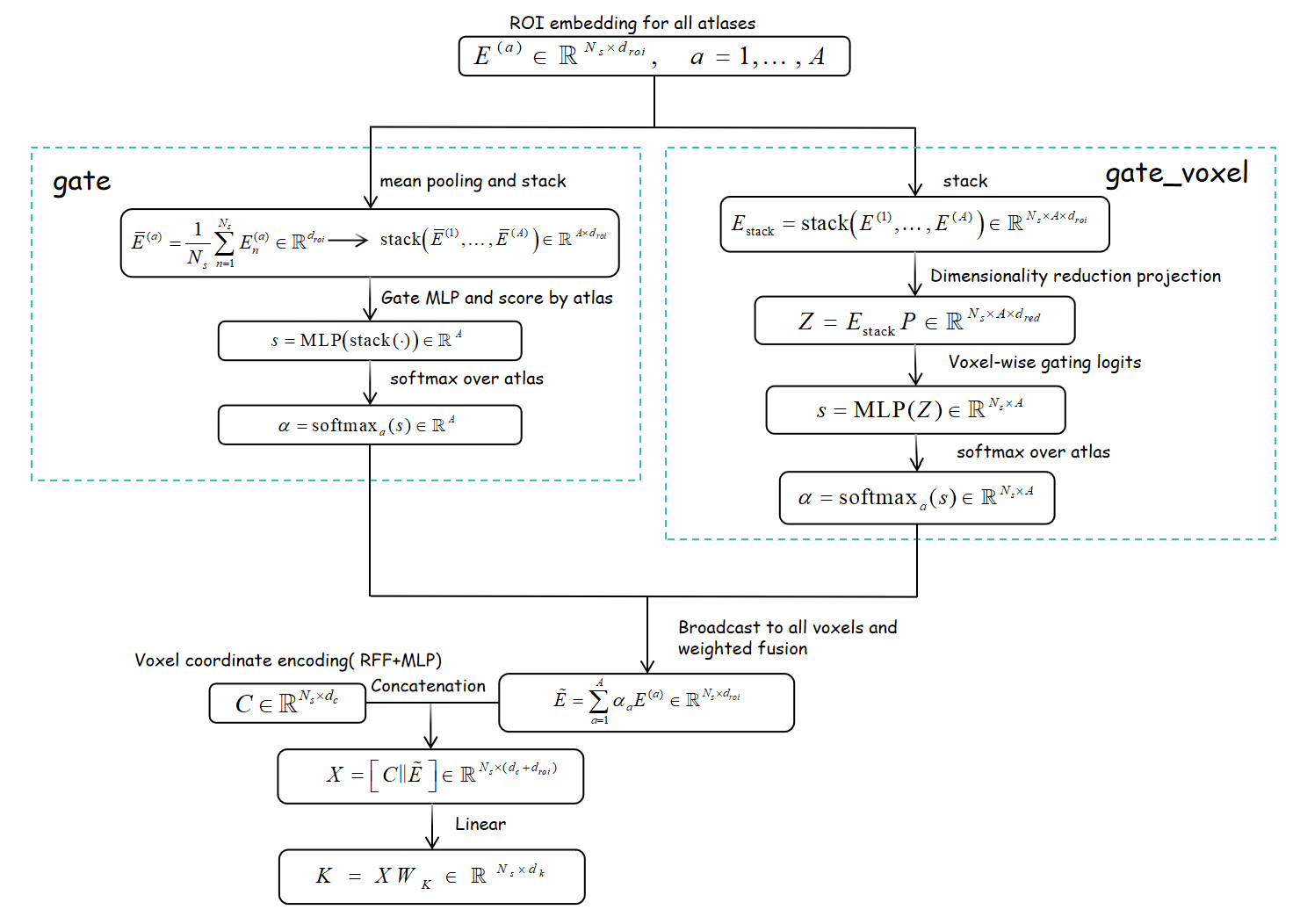}
    \caption{Mechanism diagram of the Gate and Voxel-gate strategies.}
    \label{fig:combination}
\end{figure}

As shown in Fig.~\ref{fig:combination}, in the Gate strategy branch, we first apply average pooling over the voxel dimension to the ROI embeddings from each atlas, yielding an atlas-specific global summary vector of dimension $d_{\mathrm{roi}}$. All atlas summaries are then stacked to form an atlas-level representation of size $A \times d_{\mathrm{roi}}$. Next, a gating multi-layer perceptron produces a score for each atlas summary vector (with shared parameters across atlases), and a softmax over the atlas dimension yields the global weights $\boldsymbol{\alpha}$. These weights are shared across all voxels of the same sample, so the ROI embeddings from multiple atlases can be fused using the same set of weights, resulting in a fused atlas-level ROI representation of size $N_s \times d_{\mathrm{roi}}$. Finally, the voxel coordinate encoding $C$ (of size $N_s \times d_c$) is concatenated with the fused ROI representation along the feature dimension to form $X \in \mathbb{R}^{N_s \times (d_c + d_{\mathrm{roi}})}$, which is then linearly projected to obtain the keys $K$. This strategy offers good interpretability and low computational overhead; however, since $\boldsymbol{\alpha}$ is shared across voxels, its ability to adapt to voxel-wise local variations is limited.

\paragraph{(iii) {Voxel-gate}: voxel-level gated weighting.}

As shown in Fig.~\ref{fig:combination}, in the Voxel-gate strategy branch, we first stack the ROI representations from each atlas along the atlas dimension, forming a multi-atlas feature set for each voxel. Next, a dimensionality-reduction projection is applied to the stacked ROI features to reduce the number of parameters and computational cost in the subsequent gating step. Then, a gating network assigns a score to each voxel's representation from every atlas, and a softmax over the atlas dimension produces the voxel-wise atlas weight vector $\alpha$. After broadcasting these weights along the ROI feature dimension, the voxel-level ROI representations from different atlases are fused via a weighted sum, yielding the fused voxel-level ROI representation $\tilde{E}$. Finally, the voxel coordinate encoding $C$ is concatenated with $\tilde{E}$ along the feature dimension to form $X$, which is then linearly projected to obtain the keys $K$. Compared with Gate, Voxel-gate allows the weights $\alpha_{n,a}$ to vary across voxel locations. This enables finer-grained and more adaptive fusion of multi-atlas priors at the voxel level. In practice, it often yields stronger cross-subject robustness and better local discriminability. The trade-off is extra parameters and computation introduced by the additional projection and the voxel-wise gating network.
\subsection{Interpretable Prompt Optimization}
\label{subsec:interpretable_prompt_optimization}
To address the lack of optimality in handcrafted prompts and the limited interpretability of soft/prefix prompts, we introduce an IPO-style interpretable prompt optimization loop. We first initialize a prompt pool with one or more seed prompts and maintain a fixed-size candidate set. In each iteration, we use a Qwen2.5-32B-Instruct model to generate several new prompts that differ semantically but share the same objective. We then feed the "new candidates + existing pool prompts" into the same standard inference pipeline and perform generation and scoring on a held-out few-shot validation set (BrainROI first produces the visual tokens. Then we combine these tokens with each candidate prompt and feed them into the frozen MLLM to generate a caption, which is scored against the reference captions). Finally, we rank all candidates by their scores, keep the top-$N$ prompts to update the pool, and proceed to the next iteration. After three rounds, we save the best-performing prompt as the fixed "optimal prompt" for subsequent validation and testing, while retaining the complete "prompt--score" evolution trace to ensure auditability and reproducibility. Appendix~\ref{app:ipo} details the IPO procedure and provides an example optimization trace.

\subsection{Constrained Generation and Consistency}
\label{subsec:constrained_generation}

During inference, we use beam search as the core decoding strategy and introduce parameterized constraints for repetition suppression and length preference within this framework, resulting in more stable outputs.

\paragraph{Stage 1: Beam search for candidate generation.}
During inference, the model maintains multiple high-probability partial sequences in parallel via beam search. We control the beam width by num\_beams. To reduce repetition and redundancy, we incorporate the following mechanisms during decoding:
\begin{itemize}
  \item \textbf{no\_repeat\_ngram\_size}: suppresses repeated $n$-grams during generation;
  \item \textbf{length\_penalty}: adjusts the scoring preference over candidate lengths, which helps mitigate the length bias that can arise when beam search maximizes probability.
\end{itemize}

\paragraph{Stage 2: Final sequence selection.}
After beam search terminates, the candidate sequences are sorted by the internal model score (like the length normalization in length\_penalty score). We select the top-1 sequence as the final output.

\section{Experiments and Results}
\label{sec:experiments}

\subsection{Experimental Setup}
\label{subsec:experimental_setup}

\paragraph{Dataset and evaluation protocol.}
We conduct all experiments on the NSD. A brief overview of NSD is provided in Appendix~\ref{app:nsd}. We strictly follow the standard protocol used in visual decoding studies to ensure reproducibility and fair comparison.
\begin{itemize}
  \item \textbf{Subjects and data split.} We use the standard split of four subjects who completed all scanning sessions: S1, S2, S5, and S7. For each subject, the test set contains 982 independent fMRI samples (each stimulus image is repeated three times and averaged). The remaining samples are used for training and validation.
  \item \textbf{Task and metrics.} The core task is brain captioning. We report BLEU-1/2/3/4, ROUGE-L, and CIDEr. We also report CLIP-S and RefCLIP-S as semantic consistency metrics.
\end{itemize}

\paragraph{Training setup.}
We train only the parameters in the fMRI branch, while keeping the image branch and the language model frozen and setting the random seed to 42. We use the AdamW optimizer with weight decay $0.01$ and default coefficients $\beta_1=0.9$ and $\beta_2=0.95$. We adopt a One-Cycle learning-rate schedule. Training is performed in two stages:
\begin{itemize}
  \item \textbf{Stage 1.} We train for 180 epochs with batch size 32 and a maximum learning rate of $3\times 10^{-4}$. We use mean squared error (MSE) as the reconstruction loss. We enable dropout warm-up for the first 3 epochs (attn\_dropout=0.50, ffn\_dropout=0.15).
  \item \textbf{Stage 2.} Starting from the best.pth from Stage 1, we train for another 200 epochs. We reduce the maximum learning rate to $1\times 10^{-4}$. We disable augmentation and warm-up, and keep other settings unchanged. We select the final checkpoint by the macro-average validation loss across subjects.
\end{itemize}

\subsection{Comparison with Existing Methods}
\label{subsec:comparison_existing}

To systematically validate the effectiveness of our method, we report results under two training settings. BrainROI denotes a fully parameter-shared model trained jointly on the training sets of four subjects (S1, S2, S5, and S7) and evaluated on the S1 test set. BrainROI-S1 denotes a model trained only on the training data of subject S1 and evaluated on the S1 test set. Other published baseline results in Table~\ref{tab:compare_methods} are taken from the corresponding original papers, and our method achieves higher scores on multiple metrics.

\begin{table}[t]
\centering
\caption{Comparison of brain captioning performance across methods under single-subject and cross-subject settings.}
\label{tab:compare_methods}
\resizebox{\linewidth}{!}{%
\begin{tabular}{|l|l|l|l|l|l|l|l|l|l|}
\hline
Method & BLEU-1 & BLEU-2 & BLEU-3 & BLEU-4 & ROUGE-L & CIDEr & CLIP-S & RefCLIP-S\\
\hline
BrainROI        & \textbf{0.6638} & \textbf{0.5077} & \textbf{0.3863} & \textbf{0.2911} & \textbf{0.4890} & \textbf{0.6952} & 0.5962 & \textbf{0.8069} \\
BrainROI-S1     & 0.6310 & 0.4609 & 0.3351 & 0.2431 & 0.4568 & 0.5983 & 0.5952 & 0.8027 \\
UMBRAE-S1       & 0.5763 & 0.3802 & 0.2500 & 0.1676 & 0.4215 & 0.5193 & 0.6644 & 0.7212 \\
UMBRAE          & 0.5944 & 0.4048 & 0.2766 & 0.1903 & 0.4371 & 0.6106 & 0.6778 & 0.7354 \\
MINDLLM         & 0.6175 & 0.4284 & 0.2986 & 0.2124 & 0.4582 & 0.6097 & -- & -- \\
SDRecon-S1~\cite{takagi2023}  & 0.3621 & 0.1711 & 0.0772 & 0.0343 & 0.2513 & 0.1383 & 0.6107 & 0.6636 \\
OneLLM-S1       & 0.4704 & 0.2697 & 0.1549 & 0.0951 & 0.3505 & 0.2299 & 0.5480 & 0.6128 \\
BrainCap-S1~\cite{ferrante2024} & 0.5596 & 0.3621 & 0.2270 & 0.1451 & 0.4069 & 0.4130 & 0.6431 & 0.6990 \\
VINDEX-S1~\cite{xia2025visualspace}    & 0.5799 & 0.3900 & 0.2604 & 0.1783 & 0.4227 & 0.5396 & 0.6694 & 0.7264 \\
VINDEX          & 0.6000 & 0.4072 & 0.2757 & 0.1891 & 0.4378 & 0.6032 & \textbf{0.6826} & 0.7388 \\
MindEye2~\cite{mindeye2}       & 0.5482 & 0.3860 & 0.2649 & 0.1816 & 0.4377 & 0.5570 & 0.6754 & 0.7373 \\
\hline
\end{tabular}%
}
\end{table}

Table~\ref{tab:compare_methods} summarizes brain captioning results across methods under both single-subject and cross-subject settings. Overall, BrainROI achieves the best performance on key text-generation metrics. In particular, BLEU-4 reaches 0.2911, and CIDEr reaches 0.6952. Both are clearly higher than representative cross-subject baselines. For example, compared with MINDLLM (BLEU-4 $=0.2124$), BrainROI improves BLEU-4 by $0.0787$. Compared with UMBRAE (CIDEr $=0.6106$), BrainROI improves CIDEr by $0.0846$. Our method also stays strong on recall-oriented matching and reference consistency. ROUGE-L is 0.4890, which is higher than MINDLLM (0.4582). RefCLIP-S reaches 0.8069, which is the highest in the table. It exceeds the strong cross-subject baseline VINDEX (0.7388) by $0.0681$. These results suggest that the proposed voxel-wise gated multi-atlas fusion improves generation quality and reference consistency in a stable manner.

Within our method, cross-subject training brings consistent gains over single-subject training. Compared with BrainROI-S1, BrainROI improves BLEU-4 from 0.2431 to 0.2911 (+$0.0480$) and improves CIDEr from 0.5983 to 0.6952 (+$0.0969$). This trend suggests that multi-subject data helps the model learn more generalizable neural representations. Under the single-subject setting, our model also outperforms multiple single-subject baselines on text-generation metrics. For instance, compared BrainROI-S1 with UMBRAE-S1 (BLEU-4 $=0.1676$, CIDEr $=0.5193$) and VINDEX-S1 (BLEU-4 $=0.1783$, CIDEr $=0.5396$), our model achieves clearly higher BLEU-4 (0.2431) and CIDEr (0.5983). This result supports the effectiveness of the encoder design itself.

We also observe that some methods show different preferences on CLIP-S and RefCLIP-S. For example, BrainROI has CLIP-S $=0.5962$ and RefCLIP-S $=0.8069$, while VINDEX shows CLIP-S $=0.6826$ and RefCLIP-S $=0.7388$. This difference indicates that our optimization strategy tends to guide the model toward captions that are more consistent with human references, rather than only maximizing global semantic similarity to the image. We further provide qualitative reference--candidate caption comparisons in Appendix~\ref{app:qual}.

\subsection{Ablation Studies}
\label{subsec:ablation}

\subsubsection{Ablation on Cross-subject Fusion Strategy}
\label{subsubsec:ablation_fusion}

To systematically verify the effectiveness of the proposed gated fusion mechanism, we perform a controlled ablation experiment. We keep all training, inference, and prompt settings unchanged and only modify the fusion\_mode parameter in the fMRI encoder. We compared three strategies—Concatenation, Gate, and Voxel-gate. The ablation results are reported in Table~\ref{tab:ablation_fusion}.

\begin{table}[t]
\centering
\caption{Ablation results of cross-subject fusion strategies, including Concatenation, Gate, and Voxel-gate.}
\label{tab:ablation_fusion}
\resizebox{\linewidth}{!}{%
\begin{tabular}{|l|c|c|c|c|c|c|c|c|}
\hline
Fusion mode & BLEU-1 & BLEU-2 & BLEU-3 & BLEU-4 & ROUGE-L & CIDEr & CLIP-S & RefCLIP-S \\
\hline
Voxel-gate & \textbf{0.6638} & \textbf{0.5077} & \textbf{0.3863} & \textbf{0.2911} & \textbf{0.4890} & \textbf{0.6952} & \textbf{0.5962} & 0.8069 \\
Gate        & 0.6578 & 0.4934 & 0.3771 & 0.2889 & 0.4862 & 0.6855 & 0.5956 & \textbf{0.8073} \\
Concatenation      & 0.6480 & 0.4894 & 0.3713 & 0.2837 & 0.4844 & 0.6629 & 0.5945 & 0.8041 \\
\hline
\end{tabular}%
}
\end{table}

As the results shown in Table~\ref{tab:ablation_fusion}. Voxel-gate achieves the best performance on 7 out of 8 metrics. The only exception is RefCLIP-S, where Gate is higher by a very small margin (0.0004). The advantage of Voxel-gate is more pronounced on metrics that are sensitive to higher-order $n$-gram structure and content coverage. Compared with Concatenation, Voxel-gate improves BLEU-4 from 0.2837 to 0.2911 and improves CIDEr from 0.6629 to 0.6952. It also yields consistent gains on BLEU-1/2/3, ROUGE-L, CLIP-S, and RefCLIP-S. Compared with Gate, Voxel-gate improves BLEU-4 by 0.0022 and CIDEr by 0.0097, while BLEU-1/2/3 and ROUGE-L also increase. We note that CLIP-S and RefCLIP-S are nearly tied across the three fusion modes. This suggests that their global semantic similarity is comparable. In contrast, Voxel-gate shows a stable advantage on BLEU-3/4 and CIDEr, which are more sensitive to fine-grained structure and content consistency. These results indicate that Voxel-gate is more effective at recovering syntactic patterns and detailed semantics.

Overall, we observe a consistent upward trend from Concatenation $\rightarrow$ Gate $\rightarrow$ Voxel-gate. Introducing gating to weight different atlases yields stable improvements. Refining the gating mechanism from global shared weights to voxel-wise weights further improves most metrics.

\subsubsection{Ablation on Constrained Decoding Strategy}
\label{subsubsec:ablation_decoding}

This section conducts an ablation study on decoding constraint parameters in the generation stage to isolate the contribution of the constraint mechanisms themselves. Our design focuses on two types of constraints—repetition suppression and length-preference adjustment. We include Greedy decoding as a search-strategy baseline, and take Beam search without constraints (Beam Only) as the reference setting upon which different constraints are incrementally added. All other inference hyperparameters (e.g., max\_new\_tokens) are kept identical across all settings. We construct five decoding configurations, as shown in Table~\ref{tab:decoding_settings}, and report the results in Table~\ref{tab:decoding_results}.

\begin{table}[t]
\centering
\caption{Ablation settings for constrained decoding, focusing on repetition suppression and length-preference adjustment.}
\label{tab:decoding_settings}

\begin{tabular}{|l|c|c|c|}
\hline
Method & num\_beams & no\_repeat\_ngram\_size & length\_penalty \\
\hline
Greedy & 1 & 0 & 1.0 \\
Beam Only & 6 & 0 & 1.0 \\
Beam + no\_repeat & 6 & 3 & 1.0 \\
Beam + length\_penalty & 6 & 0 & 0.1 \\
Full constraints & 6 & 3 & 0.1 \\
\hline
\end{tabular}%

\end{table}

\begin{table}[t]
\centering
\caption{Ablation results for constrained decoding strategies, including repetition suppression and length-preference adjustment.}
\label{tab:decoding_results}
\resizebox{\linewidth}{!}{%
\begin{tabular}{|l|c|c|c|c|c|c|c|c|}
\hline
Method & BLEU-1 & BLEU-2 & BLEU-3 & BLEU-4 & ROUGE-L & CIDEr & CLIP-S & RefCLIP-S \\
\hline
Greedy & 0.5342 & 0.3336 & 0.2158 & 0.1450 & 0.3867 & 0.3914 & 0.5874 & 0.7831 \\
Beam Only & 0.6300 & 0.4634 & 0.3432 & 0.2552 & 0.4632 & 0.5945 & 0.5908 & 0.7967 \\
Beam + no\_repeat & 0.6304 & 0.4644 & 0.3440 & 0.2556 & 0.4636 & 0.5947 & 0.5907 & 0.7968 \\
Beam + length\_penalty & 0.6309 & 0.4650 & 0.3448 & 0.2561 & 0.4641 & 0.5950 & 0.5908 & 0.7970 \\
Full constraints & \textbf{0.6638} & \textbf{0.5077} & \textbf{0.3863} & \textbf{0.2911} & \textbf{0.4890} & \textbf{0.6952} & \textbf{0.5962} & \textbf{0.8069} \\
\hline
\end{tabular}%
}
\end{table}

As shown in Table~\ref{tab:decoding_results}, the main gain comes from the search strategy itself. Greedy yields BLEU-4 $=0.1450$ and CIDEr $=0.3914$, while Beam Only improves them to 0.2552 and 0.5945. This result shows that expanding the candidate search space substantially improves higher-order $n$-gram matching and key-information coverage, leading to more complete captions and better reference consistency. RefCLIP-S also increases from 0.7831 to 0.7967, which suggests improved semantic consistency, although the margin is smaller.

On top of beam search, a single constraint produces only minor changes. Adding no\_repeat slightly improves BLEU-4 and CIDEr (0.2552$\rightarrow$0.2556 and 0.5945$\rightarrow$0.5947). This gain may come from reduced local $n$-gram redundancy, which allows the model to include more informative content and thus improves higher-order matching and content coverage. Adding length\_penalty also gives a small improvement (BLEU-4 $=0.2561$, CIDEr $=0.5950$), indicating that length preference can mitigate length bias or redundancy. However, using it alone is still insufficient to substantially change overall generation quality.

The combined setting (Full constraints) achieves the best results across all metrics. BLEU-4, CIDEr, and RefCLIP-S increase to 0.2911, 0.6952, and 0.8069, respectively. Compared with Beam Only, the CIDEr gain is more pronounced. This suggests that the combination of constraints better increases the coverage of key information and reduces unhelpful repetition. The simultaneous increase in RefCLIP-S indicates that the improvement is not limited to surface-level overlap but also reflects better consistency with visual semantics.

\subsubsection{Subject-specific evaluation under cross-subject joint training (S1, S2, S5, S7).}
\label{subsec:subject_specific_eval}

\begin{table}[t]
\centering
\caption{Subject-specific evaluation under cross-subject joint training.}
\label{tab:subject_specific}
\resizebox{\linewidth}{!}{%
\begin{tabular}{|l|c|c|c|c|c|c|c|c|}
\hline
Subject & BLEU-1 & BLEU-2 & BLEU-3 & BLEU-4 & ROUGE-L & CIDEr & CLIP-S & RefCLIP-S \\
\hline
S1 & \textbf{0.6638} & \textbf{0.5077} & \textbf{0.3863} & \textbf{0.2911} & \textbf{0.4890} & \textbf{0.6952} & \textbf{0.5962} & \textbf{0.8069} \\
S2 & 0.6352 & 0.4754 & 0.3572 & 0.2691 & 0.4676 & 0.6128 & 0.5906 & 0.7955 \\
S5 & 0.6414 & 0.4749 & 0.3500 & 0.2593 & 0.4696 & 0.6178 & 0.5920 & 0.7996 \\
S7 & 0.6130 & 0.4494 & 0.3307 & 0.2473 & 0.4538 & 0.5392 & 0.5871 & 0.7874 \\
\hline
\end{tabular}%
}
\end{table}

In this section, we provide additional ablation results for the subject-specific evaluation under cross-subject joint training. We jointly train a single fully parameter-shared model on the training sets of S1, S2, S5, and S7, and then evaluate it on each subject's own test set. The results are shown in Table~\ref{tab:subject_specific}.

As shown in Table~\ref{tab:subject_specific}, under the joint-training and subject-wise testing setting, S1 achieves higher scores than S2/S5/S7 on key metrics. For example, BLEU-4 is higher by 0.0220/0.0318/0.0438, and CIDEr is higher by 0.0824/0.0774/0.1560, respectively. In contrast, the differences in semantic similarity are small. CLIP-S and RefCLIP-S only drop slightly across subjects.

S7 shows the lowest performance (BLEU-4 $=0.2473$, CIDEr $=0.5392$). The gap is more evident on fine-grained alignment-related metrics (BLEU-4 is lower by 0.0438 and CIDEr is lower by 0.1560 compared with S1). However, the fluctuations in semantic similarity remain small (CLIP-S is lower by only 0.0091, and RefCLIP-S is lower by only 0.0195). This pattern suggests that cross-subject differences mainly affect readability and fine-grained $n$-gram matching rather than overall semantics.

S2 and S5 perform similarly. S2 has a slightly higher BLEU-4 (0.2691), while S5 has a slightly higher CIDEr (0.6178). Overall, the results show a stable ordering of \mbox{S1 $>$ S2 $\approx$ S5 $>$ S7}. The differences are mainly reflected in BLEU-4 and CIDEr, while CLIP-S and RefCLIP-S vary within a relatively narrow range.

\section{Limitations and Future Work}
\label{sec:limitations_future}

\paragraph{(1) Data and modality coverage.}
Our current evaluation focuses on static fMRI signals and image-level semantics. We do not yet cover videos, speech, or broader multimodal signals such as MEG/EEG. In future work, we plan to extend the framework to temporal modeling and multimodal fusion. We will also systematically study generalization across tasks and across decoding paradigms.

\paragraph{(2) Evaluation and reliability.}
Our current results mainly rely on traditional automatic metrics such as BLEU. These $n$-gram overlap-based measures do not adequately capture semantic consistency, factual correctness, readability, or diversity. They can undervalue semantically equivalent paraphrases and may be less robust for longer texts. Future work should adopt evaluation protocols that better reflect semantic alignment and factual correctness.

\section{Conclusion}
\label{sec:conclusion}

This paper introduces a subject-agnostic encoder inspired by MINDLLM and a multi-atlas soft-ROI alignment scheme for cross-subject modeling. We further propose voxel-wise Voxel-gate fusion to robustly integrate multi-atlas priors. In addition, we incorporate Qwen-based interpretable prompt optimization and apply constrained decoding at inference, including beam search, repetition suppression, and length penalty. As a result, our final model achieves consistent improvements over prior methods across multiple text-based metrics.

\clearpage
\section*{Appendix}
\appendix
\renewcommand{\theHsection}{app.\Alph{section}}

\section{Natural Scenes Dataset (NSD) Overview}\label{app:nsd}
The NSD is a large-scale natural-scene fMRI benchmark for visual neuroscience and NeuroAI. NSD was acquired with whole-brain ultra-high-field 7T fMRI (1.8\,mm isotropic resolution; TR = 1.6\,s) and, through a multi-session longitudinal design, provides high-volume, high-SNR visually evoked responses for each individual subject. Data collection typically spans 30--40 scan sessions (distributed weekly over roughly one year). Each subject views approximately 9,000--10,000 distinct color natural-scene images, totaling about 22,500--30,000 stimulus presentations. The stimuli are mainly drawn from natural image repositories such as MS COCO, covering diverse objects, scenes, and compositional relationships, which offer substantial semantic diversity and difficulty for tasks such as predicting visual semantics or descriptive text from brain signals.

In addition to the main-task trials, NSD provides various auxiliary data and preprocessed derivatives, including (but not limited to) high-resolution structural scans (e.g., T1/T2), functional localizers and retinotopic mapping (retinotopy) as well as category localizers, resting-state data, physiological recordings, and partial eye-tracking. NSD also supplies stimulus-aligned indexing information and preprocessed voxel-wise response representations, facilitating the development of end-to-end decoding models and cross-subject alignment methods.

A key feature of NSD is that it contains both subject-specific and cross-subject shared stimulus sets. Under commonly used cross-subject evaluation protocols, subjects who completed all scan sessions are selected, and a set of images viewed by all subjects is used as a shared test set (we use 982 shared images in this work). Beyond the shared set, the remaining images are largely mutually exclusive across subjects and are used for training and validation. Because subjects differ in the number of available voxels and spatial coverage, and because the shared test set is relatively small, this setting naturally emphasizes cross-subject generalization and alignment, which also motivates our cross-subject soft-ROI fusion and unified representation learning strategy.

\section{IPO and an Example Trace}\label{app:ipo}

\begin{figure}
\includegraphics[width=\textwidth]{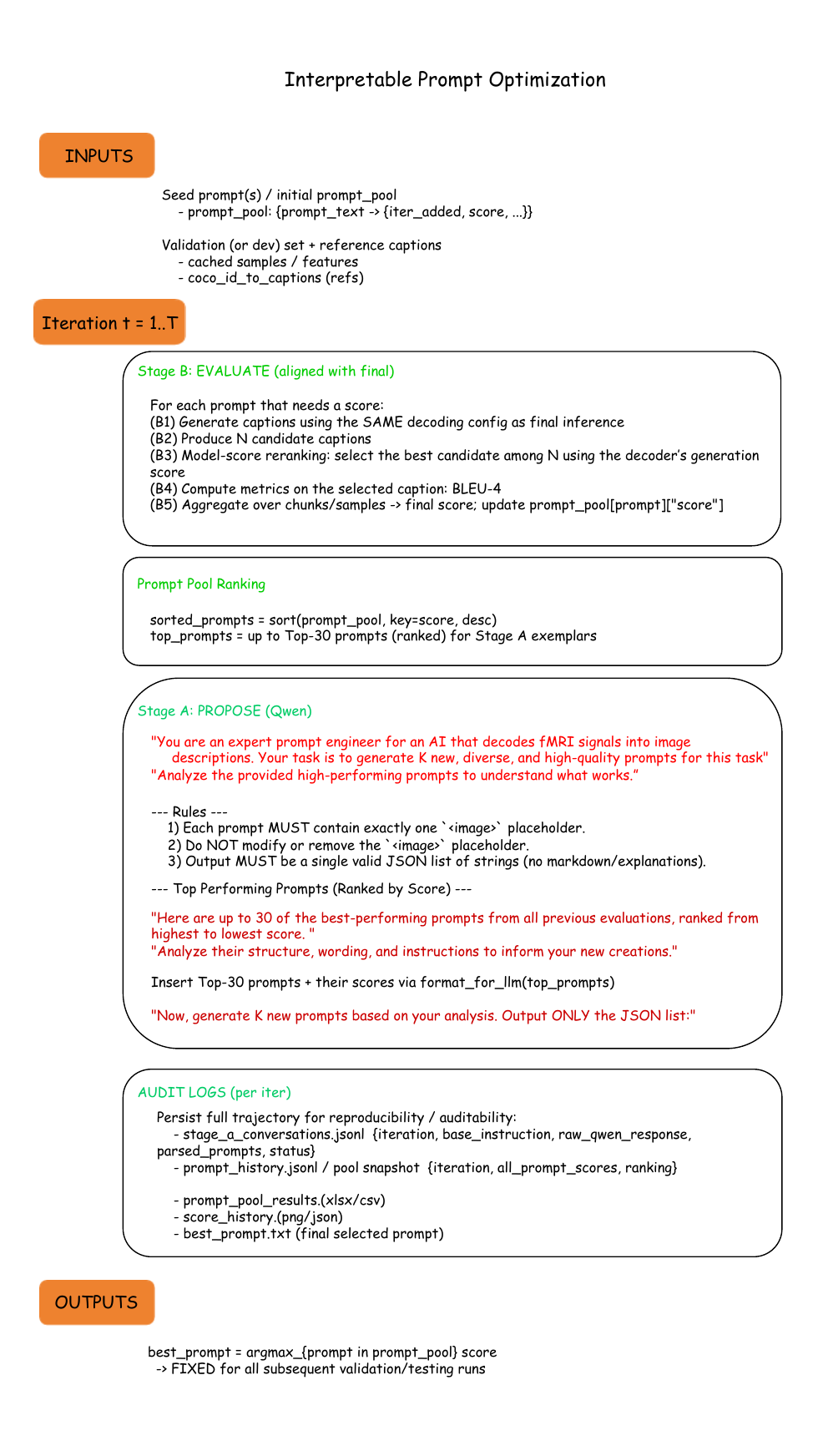}
\caption{Interpretable Prompt Optimization and an Example Trace.} \label{IPO}
\end{figure}

\begin{table}[t]
\centering
\caption{Example IPO trace ranked by validation BLEU-4 score.}
\label{tab:ipo_trace}
\resizebox{\linewidth}{!}{%
\begin{tabular}{|l|c|c|}
\hline
Prompt & iter\_added & BLEU-4 \\
\hline
What is the content of the image \texttt{<image>}? Please answer in short sentences. & 3 & 0.2911 \\
Describe this image \texttt{<image>} as simply as possible. & 3 & 0.2910 \\
Summarize the content of the photo \texttt{<image>}. & 3 & 0.2899 \\
Outline the major subjects and background details within \texttt{<image>}. & 3 & 0.2895 \\
Identify and explain the key visual components in the scene of \texttt{<image>}. & 2 & 0.2866 \\
Analyze the composition and focal points of the scene captured in \texttt{<image>}. & 3 & 0.2865 \\
Outline the central features and mood conveyed in \texttt{<image>}. & 2 & 0.2820 \\
What elements are present in the scene depicted by \texttt{<image>}? & 1 & 0.2767 \\
Identify and explain the main components in the visual content of \texttt{<image>}. & 2 & 0.2735 \\
Identify and describe the main action taking place in \texttt{<image>}. & 1 & 0.2630 \\
\hline
\end{tabular}%
}
\end{table}

This section supplements the main text by describing the reproducible scheme for interpretable prompt optimization. We view IPO as a closed-loop search process of "propose–evaluate–select," which automatically explores human-readable prompts on a small validation set and selects candidate prompts using a fixed metric. We use BLEU-4 as the ranking metric: BLEU-4 emphasizes higher-order n-gram phrase-level matching, which is more sensitive to wording drift induced by prompts, and can suppress "superficial fitting" that only improves unigram overlap. We notice that the ranking trends of BLEU-4 are largely consistent with those of CIDEr, ROUGE-L, and CLIP-S/RefCLIP-S; therefore, we adopt BLEU-4 as a proxy ranking metric during prompt search.

We use a locally deployed Qwen-2.5-32B-Instruct model to generate candidate prompts (temperature = 0.8, top-p = 0.95, max\_new\_tokens = 1024). The initial prompt\_pool size is 5; at each iteration, 6 new prompts are generated, and the optimization runs for 3 iterations. To ensure reproducibility and auditability of the prompt optimization process, we fully record at each iteration the seed prompts, the model-generated candidate prompts, the corresponding validation scores, and the ranking results, and we fix the finally selected best\_prompt for all subsequent validation and testing. Fig.~\ref{IPO} presents the complete IPO pipeline, and Table~\ref{tab:ipo_trace} shows an example trace ranked by BLEU-4 (each row corresponds to a candidate prompt). Here, iter\_added denotes the IPO iteration in which a prompt is first added to prompt\_pool, and BLEU-4 denotes the Stage B validation score.

\section{Qualitative Examples of Brain Captioning}\label{app:qual}

\begin{figure}
    \centering
    \includegraphics[width=1\linewidth]{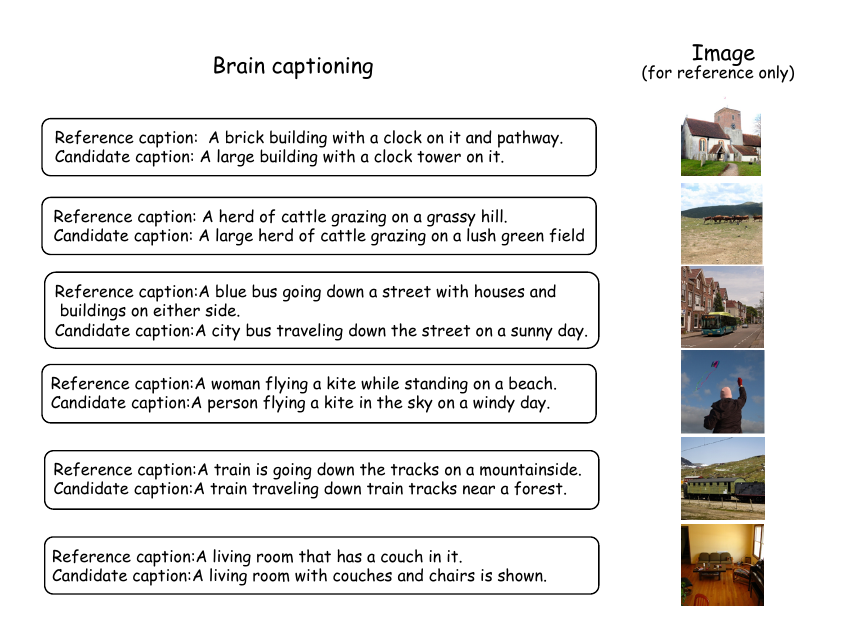}
    \caption{Qualitative comparison of COCO reference captions and fMRI-generated candidate captions.}
    \label{fig:qualitative_pairs}
\end{figure}

This section provides qualitative examples to complement the quantitative results. Figure~\ref{fig:qualitative_pairs} presents representative reference--candidate caption pairs. For each stimulus, we report the COCO reference caption and the caption generated from fMRI by our method; the corresponding image is shown for visualization. Overall, the generated captions often capture the key entities and actions consistent with the references, while most differences arise from fine-grained attributes or background details.

%
%
%
%

\FloatBarrier

\end{document}